# A BERT-Based Summarization approach for depression detection

Hossein Salahshoor Gavalan[1], Mohmmad Naim Rastgoo[2], Bahareh Nakisa[2]

[1] Faculty of Science, University of Tehran, Tehran, Iran
[2] Faculty of Science Engineering and Built Environment, School of Information Technology, Deakin University, Vic, Australia

***Abstract* – Depression is a globally prevalent mental disorder with potentially severe repercussions if not addressed, especially in individuals with recurrent episodes. Prior research has shown that early intervention has the potential to mitigate or alleviate symptoms of depression. However, implementing such interventions in a real-world setting may pose considerable challenges. A promising strategy involves leveraging machine learning and artificial intelligence to autonomously detect depression indicators from diverse data sources. One of the most widely available and informative data sources is text, which can reveal a person's mood, thoughts, and feelings. In this context, virtual agents programmed to conduct interviews using clinically validated questionnaires, such as those found in the DAIC-WOZ dataset, offer a robust means for depression detection through linguistic analysis. Utilizing BERT-based models, which are powerful and versatile yet use fewer resources than contemporary large language models, to convert text into numerical representations significantly enhances the precision of depression diagnosis. These models adeptly capture complex semantic and syntactic nuances, improving the detection accuracy of depressive symptoms. Given the inherent limitations of these models concerning text length, our study proposes text summarization as a preprocessing technique to diminish the length and intricacies of input texts. Implementing this method within our uniquely developed framework for feature extraction and classification yielded an F1-score of 0.67 on the test set—surpassing all prior benchmarks—and 0.81 on the validation set—exceeding most previous results—on the DAIC-WOZ dataset. Furthermore, we have devised a depression lexicon to assess summary quality and relevance. This lexicon constitutes a valuable asset for ongoing research in depression detection.**

## 1. Introduction

Depression is a prevalent type of mood disorder. It affects approximately 280 million individuals worldwide which can cause the affected person to function poorly at work, school, and within the family. In the worst situations, depression may trigger suicidal thoughts and behaviors. Over 700,000 individuals every year pass away from severe depression, according to data from the World Health Organization (WHO) [1].

Machine learning and AI algorithms are increasingly being utilized as computer-assisted tools to support clinicians and professionals, empowering them in the detection of mental diseases. AI can play a vital role in detecting depression and providing early intervention for people who suffer from this mental health condition. Although according to earlier research, depressive symptoms may be prevented by intervening during the first depressive episode [2], early illness intervention, meanwhile, might be challenging due to some barriers such as stigma [3, 4], prejudice, stereotyped ideas, and discriminatory actions towards the depressed person [5]. As a consequence, psychologists are unable to effectively intervene or even accurately gauge the degree of depression. In light of this, AI can help overcome some of the barriers by analyzing various signals from speech [6], facial expressions [7], text [8], and behavior to identify signs of depression and provide feedback, support, or referrals to appropriate professionals. AI can also monitor the progress of treatment and alert clinicians if there are any changes or concerns. Hence, early intervention may be accomplished by giving patients and mental health providers access to an objective depression detection system.

There are different datasets for depression detection. One of the public ones in depression detection is the Distress Analysis Interview Corpus Wizard-of-Oz (DAIC-WOZ) dataset [9]. This dataset contains clinical interviews of 189 participants designed to support the diagnosis of psychological distress conditions such as anxiety, depression, and post-traumatic stress disorder. Interviews are conducted by an animated virtual interviewer called Ellie, controlled by a human interviewer in another room. A digital video camera was used to record the patient's audio and video data, and the audio was then used to create the text data.

Various modalities, such as textual, speech, and facial features of DAIC-WOZ can be used for depression detection. There are several studies [10-15] based on all of these modalities. For example, L. Yang et al. [15] used all modalities and achieved an F1-score of 0.86 on the development set and 0.57 on the test set on depression detection. Although author won the challenge of AVEC-2016, they utilized all the modalities in this dataset (text, facial features, speech) which limits the application of the model in the real-world application. There are some other studies [16-19] that only focused on text and audio and achieved acceptable performance.

Among different modalities, text inputs can be considered as a more accessible modality for depression detection. There are several reasons for this. First, text inputs do not require specialized equipment, place (like a hospital), or software, unlike speech and facial features that need microphones and cameras. Second, text inputs can be collected from various sources, such as social media posts, online forums, and chat logs, while speech and facial features are more limited in availability and privacy. Third, text inputs can capture more semantic information and diverse expressions of emotions, thoughts, and behaviors that are indicative of depression, while speech and facial features may be affected by noise, accent, speech rate, intonation, environment, or cultural factors.

Therefore, text inputs are more accessible modalities for depression detection [20, 21].

To analyze text using machine learning algorithms, it is necessary to represent the text in a numerical format, typically as a vector, that can be processed by these algorithms. Word embedding is a process of representing words as numerical vectors that capture their semantic and syntactic properties. These numerical vectors can be utilized by various deep learning or machine learning algorithms for diverse tasks.There are several techniques to generate embeddings from text such as TF-IDF, Glove [22], BERT [23], and ELMo [24]. BERT, as one of the state-of-the-art techniques in various natural language processing tasks, is based on a bidirectional transformer model that can learn from both the left and right contexts of a word. It is based on the idea of training the network on a massive corpus of unlabeled text and then fine-tuning it on a specific downstream task. Since the release of the original BERT model, several more efficient variants have been developed, such as DistilBERT and ALBERT, which aim to enhance performance and reduce computational requirements. DistilBERT [25] is a smaller and faster variant of BERT that is obtained by applying knowledge distillation techniques to the BERT model which reduces the number of layers by half and removes some components such as the token-type embeddings and the pooler. ALBERT [26] is another light version of BERT which aims to enhance the efficiency and speed of the model by minimizing its size. This model employs two primary parameter-reduction techniques, namely factorized embedding parameterization and cross-layer parameter sharing.

Despite the power of BERT-based networks, the maximum input length of 512 tokens (which restricts the amount of context that BERT-based networks can process) makes them challenging to apply to certain domains and applications. This is especially problematic for domains that deal with lengthy texts, such as clinical texts that contain important information for diagnosis and treatment. While modern large language models (LLMs) can effectively process extensive contextual data, however, their practical deployment faces significant challenges. Firstly, in scenarios where sufficient labeled data for training and fine-tuning these models is lacking. Secondly, the deployment of such models is constrained by the high cost of hardware capable of running them.

The DAIC-WOZ dataset as one of the unique datasets for depression detection, contains numerous documents that exceed the mentioned token limitation of BERT-based models. In light of the growing length of textual documents, particularly in clinical contexts, and the robustness of BERT-based networks, it is imperative to use suitable methodologies to address this obstacle. One approach in some studies [18, 21] is to use a sequence setting in which a BERT encoder receives sentences one by one and feeds them to a BiLSTM for classification. However, the sequential arrangement of this method limits the ability of the BERT network to capture the overall sentiment of the text. An alternative approach in some papers [27, 28] is to split the document into chunks that do not exceed the context window size and feed them to BERT networks to extract the corresponding embeddings. The mean of all embeddings then serves as the document representation. While this method preserves contextual information better than the first approach, its drawback lies in its lack of transparency, as it is harder to explain which part of the text contributes to the final decision.

There is also a third technique, called summarization, that identifies sentences (or key phrases) that encapsulate the essence of the ideas presented in a given text. Text summarization not only shortens documents but also identifies the most relevant phrases and terms that are important for linguistic analysis. There are different ways of summarization methods that are based on statistical (RAKE [29], YAKE [30]), graph-based (TextRank [31]), and deep learning (KeyBERT [32]) approaches. N. Giarelis et al. [33] has showed that KeyBERT is a promising method that surpasses other key phrase extraction algorithms in long text summarization.

Similar to some other medical datasets, DAIC-WOZ faces the challenge of data imbalance. If not carefully addressed, this imbalance can significantly impair the performance of machine learning algorithms, leading to biased models that favor the majority class over the minority [34]. One of the approaches to deal with imbalanced text data is data resampling to create a more balanced distribution of samples by either oversampling the minority class or undersampling the majority class. There are some studies [17, 18] that applied data resampling on the minority class to overcome the data imbalanced issue in DAIC-WOZ dataset. However, data resampling has some disadvantages, such as introducing noise, distorting the original meaning and context of the text, or overfitting in the case of oversampling, and losing valuable information or underfitting in the case of undersampling [35, 36]. Because of all these reasons, it can adversely affect the generalization of the model on unseen data. Another way to address data imbalance is by using a more robust loss function. The Focal Loss function [37] addresses class imbalance during training which not only allocates varying weights to different classes depending on their frequency, but it also concentrates on cases that are difficult to categorize. This results in a more robust model by decreasing the loss contribution of simple cases and increasing the loss contribution of difficult ones.

A depression lexicon consists of words (so-called alerting words) that reflect the cognitive, emotional, and behavioral aspects of depression. These words are derived from the language use of individuals experiencing depression and serve as a significant resource for both research and therapeutic applications. Alerting words have a critical role in the early prediction of depression and, consequently, effective depression treatment. Studies such as [38-40] have constructed a depression lexicon and then utilized their lexicon to build a depression detection model using general data collected from social media. The outcome of the study demonstrates that the incorporation of a vocabulary may significantly enhance depression classification. In the case of the DAIC-WOZ dataset, there are few studies, such as those by Y. Zhang [21] and E. Villatoro-Tello [27], that used a depression lexicon for depression classification. However, neither of these two studies provided a comprehensive depression vocabulary and only presented a limited range of key phrases.

The following is the summary of our contributions:

- We propose a combinatorial and novel framework for automated depression detection from text based on text summarization techniques and BERT-based models. Our approach applies KeyBERT, a keyword extraction method, to effectively summarize preprocessed transcripts and overcome token limitations. Our method facilitates the identification and extraction of the most informative phrases, resulting in an effective reduction of the document's length to around 512 tokens. To the best of our knowledge, this is the first study to apply a summarization approach for depression detection using DAIC-WOZ dataset. This study demonstrates that our approach yields promising outcomes to assist psychologists in detecting depression more accurately at an early stage.

- In order to address the challenge of imbalanced data, we implemented Focal Loss to give greater weight to the minority class, which in this case is the group of depressed individuals. This approach not only directs more attention to the minority class, but also prioritizes hard examples, leading to improved performance.
- To test KeyBERT's ability in identifying key phrases, we developed a comprehensive lexicon for depression by extracting data from the DAIC-WOZ dataset and then compare it with the KeyBERT-generated summarized texts. We created this lexicon by implementing a new pipeline that uses a scoring system based on TF-IDF score and cosine similarity of text embeddings. This vocabulary can be effectively used and explored in future research on depression detection.

The remainder of this paper is organized as follows: Section two describes the proposed method in detail, including information related to the data. Section three discusses the experiment setup, results, discussion, and conclusions.

## 2. Proposed Method

This section provides a general overview of our proposed text-based depression detection model utilizing the DAIC-WOZ dataset. To overcome the challenge posed by lengthy documents, we applied KeyBERT for text summarization and three different BERT-based models for word embedding. In this study, we are not only proposing a novel framework for depression classification but also building a comprehensive depression lexicon using the DAIC-WOZ dataset and alerting words from prior studies on social media.

As depicted in Fig.1, there are three main blocks. The data preprocessing stage converts transcripts into numerical vectors. This is done by first removing the virtual agent's inquiries and summarizing the remaining texts using KeyBERT. Then, transforming the summaries into numerical matrices by virtue of the BERT-based encoders. These matrices are then pooled by averaging over the rows. The depression detection pipeline uses two fully connected layers and 1D Convolutional Neural Networks (CNNs) to extract features from numerical vectors. These features are used as input for training the classification layer using a suitable loss function and optimizer. The output is a binary label, where 1 indicates depression and 0 indicates no depression. The process of the depression lexicon construction is applied as follows: First, we apply a scoring system based on term frequency-inverse document frequency (TF-IDF) with some modifications to extract the most informative words from the transcripts. Second, we measure the cosine similarity between these words and a depression lexicon developed from earlier studies on depression-related language on social networks and pick words with higher score to build our own depression lexicon. A comparison step between keywords from the summarized and full texts is considered to assess the effectiveness of KeyBERT's capabilities in detecting the most representative words and phrases.

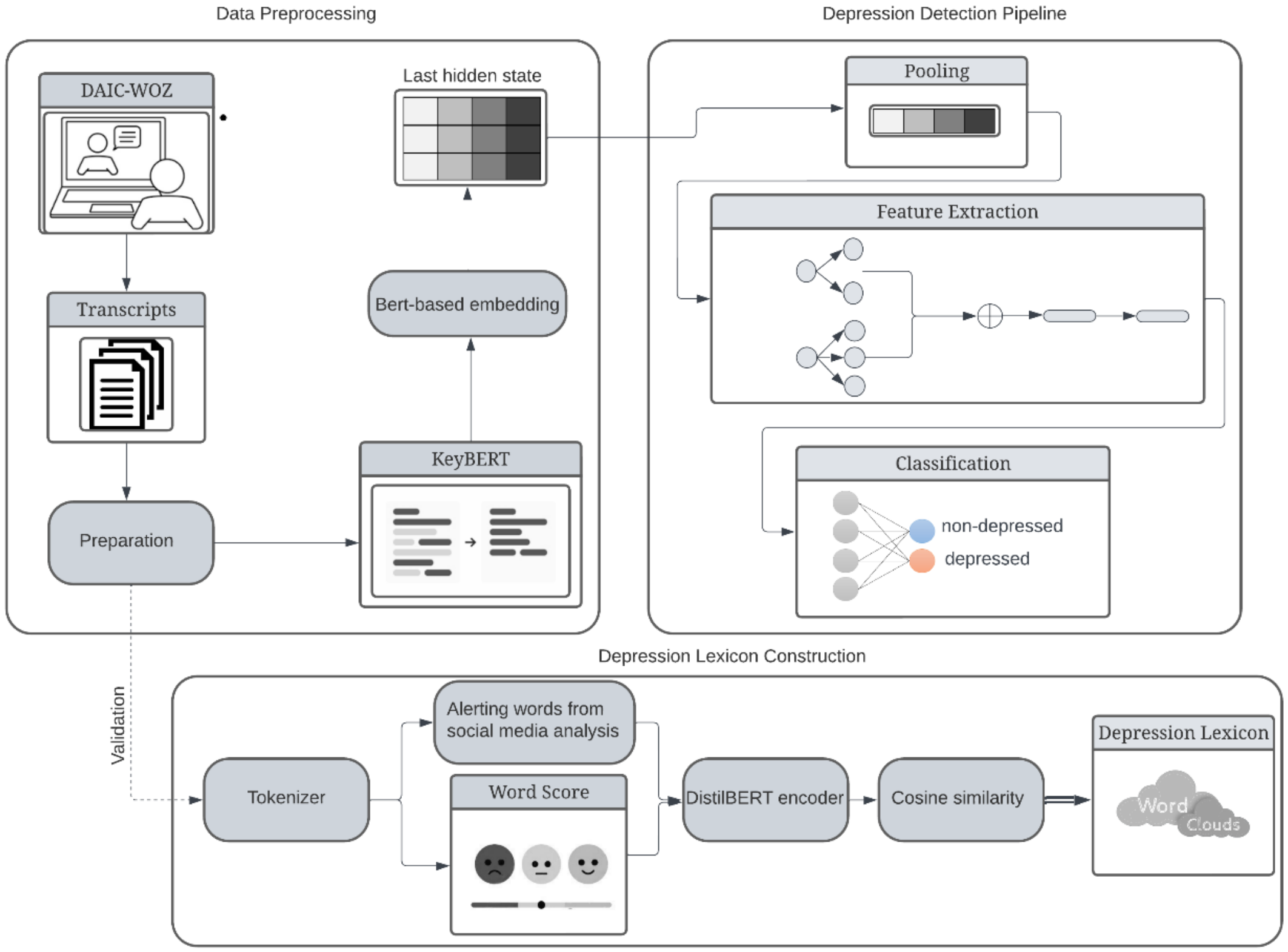

**Fig. 1.** The proposed pipeline for depression detection plus depression lexicon construction using the DAIC-WOZ dataset

### *2.1. DAIC-WOZ Dataset*

The DAIC-WOZ dataset is a collection of clinical interviews with 189 participants who suffer from psychological distress conditions such as anxiety, depression, and post-traumatic stress disorder and is publicly available on a website provided by USC's Institute of Creative Technologies (http://dcapswoz.ict.usc.edu). The interviews were conducted by Ellie, an animated virtual interviewer that was controlled by a human interviewer in another room. The Patient Health Questionnaire (PHQ-8) [41] score from the DAIC-WOZ reveals the degree of depression for each participant. In order to indicate the existence of depression, the PHQ-8 score is additionally given a binary label. If the person receives a score of more than or equal to 10, they are considered to be depressed and vice versa (PHQ score equals to zero means no symptom of depression and 21 means a high level of depression). A training set, a development (validation) set, and a test set are all included in the DAIC-WOZ dataset.

Figure 2 depicts a statistical summary of the dataset. As mentioned before, it is obvious that the DAIC-WOZ dataset is heavily imbalanced in each three of sets.

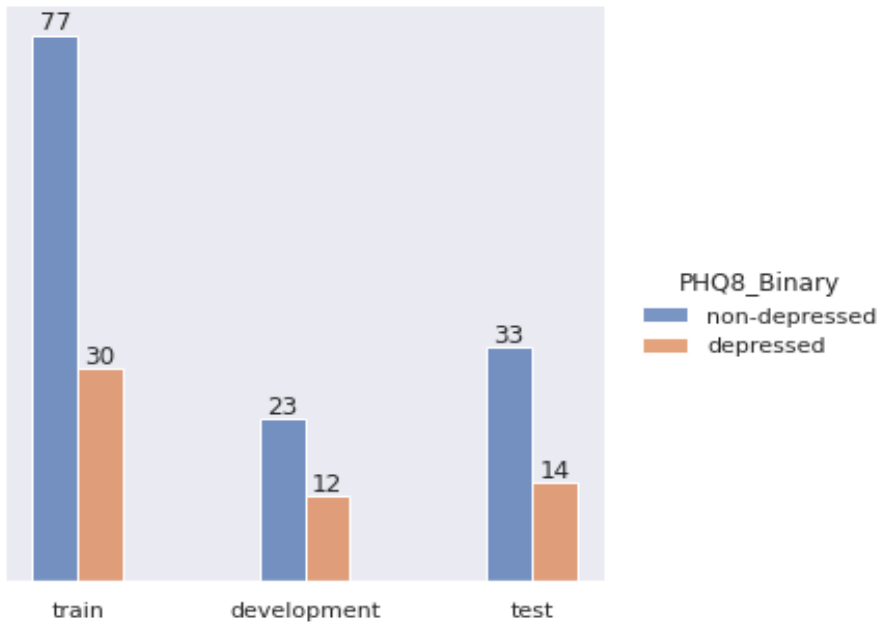

**Fig. 2.** Percentage of depressed versus non-depressed interviewers in each set of DAIC-WOZ dataset

### *2.2. Data Preprocess*

#### *2.2.1. Preparation*

Following a context-free approach, we focused solely on the responses given by the participants and disregarded Ellie's questions. According to Fig. 3, a participant's response to a question is treated as one sentence for our data analysis.

It should be noted that Transcripts for the Participants with IDs equal to 451, 458, and 480 did not have Ellie's questions. Therefore, we decided to use their voice recording to separate different answers from each other and include them in our dataset.

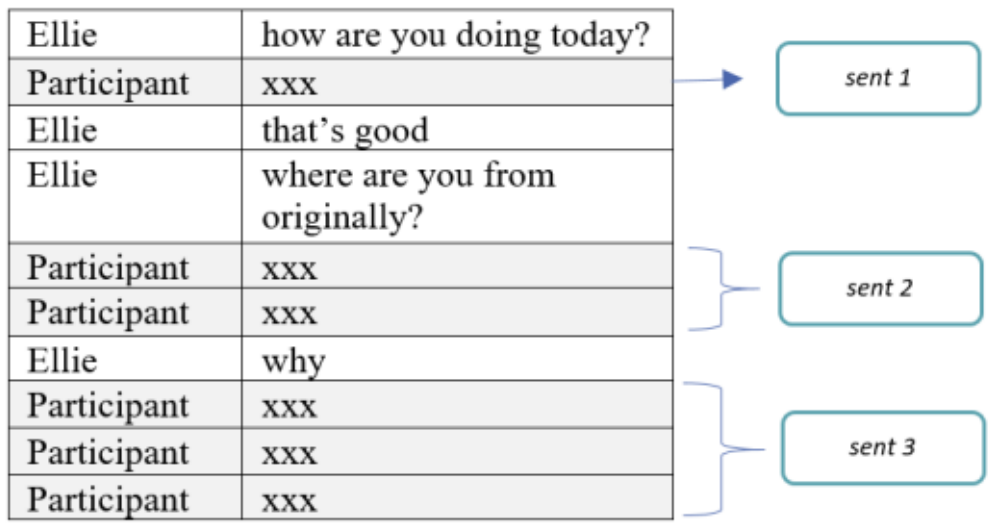

**Fig. 3.** An example of DAIC-WOZ transcripts. The whole answer of the participant to each question is regarded as a sentence.

#### *2.2.2. KeyBERT*

KeyBERT is a Python package that uses BERT models to extract keywords or key phrases from text documents and summarize them. In the context of the DAIC-WOZ dataset, a significant challenge arises due to the fact that the majority of the transcripts surpass the maximum length permitted by BERT-based models (512 tokens). Summarization is crucial in our study because it ensures compatibility with BERT model's input specifications and helps retain the most important information. This is essential for effective model training and subsequent tasks such as classification or construction of a depression lexicon.

The summarizing procedure of KeyBERT works as follows: First it generates a list of candidate n-gram key phrases for each document. Second, a document embedding vector is created, as well as an embedding vector for each potential key phrase. These embeddings are generated by utilizing pre-trained BERT-based models in the sentence-transformer package. Noting that all models have a token limit that varies from model to model, each document may need to be split into smaller chunks to fit this limit and then recombine the most informative phrases of chunks to obtain a final summarization of the document with approximately 512 tokens. The basic assumption in KeyBERT is that the more similar the vector representation of a word or phrase is to the vector representation of the document, the more likely it is to capture the main idea and topic of the document. Therefore, the third stage involves calculating the cosine similarity between the embedding vectors of candidate key phrases and the document. Afterwards, it ranks the key phrases by their similarity scores, from high to low, as a way of selecting the most relevant ones.

KeyBERT differs from other approaches in that it adds a diversification step to the output. This step uses either the Maximal Marginal Relevance [40] or Max Sum Similarity metric to diversify the key phrases. These metrics have some parameters that need to be adjusted to avoid too much similarity among the key phrases, while maintaining the model's accuracy. In this paper, to address the drawbacks of highly similar results, a diversification step is applied using the Maximal Marginal Relevance (MMR) measure. This measure is defined as follows:

$$MMR = argmax_{C_i \in C \setminus S}\left[\lambda(\widetilde{cos}_{sim}(C_i, doc) - (1 - \lambda) max_{C_j \in S} \widetilde{cos}_{sim}(C_i, C_j))\right] \quad (1)$$

where $C, S$ represent the set of (ranked) extracted key phrases and subset of documents in $C$ which are already selected respectively ( $C/S$ means set of unselected documents in $C$ ). $doc$ is the document embedding vector, $\widetilde{cos}_{sim}$ is a normalized cosine similarity function between two vectors (that are defined in the parenthesis), and λ is a parameter that determines the trade-off between relevance and diversity of the candidate key phrases. The subtraction of the diversity term (the second part of the formula) from the relevance term (the first part of the formula) ensures that the selected document is not only relevant but also provides new information, thus maintaining diversity in the results.

*2.2.3. BERT-based embedding*

In our study, we utilized three prominent language models, BERT, DistilBERT, and ALBERT, to evaluate their performance in the classification task of this specific dataset. Each model was trained on an identical dataset to ensure a fair comparison of their capabilities. Despite the shared training data, each model employs its unique tokenization and embedding methods, resulting in distinct numerical matrices.

There are several BERT-based models that have been pre-trained on a wide range of datasets. Our experiments have shown that the "base-uncased" models, which were trained using a combination of Wikipedia and BookCorpus data, delivered the best performance.

The encoder part of these models is responsible for transforming text into a numerical matrix. They receive tokenized text as input and produce a numerical matrix as the output with the shape of the batch size, sequence length, and hidden size. For all models, the hidden size and sequence length are 768 and 512, respectively. This numerical matrix captures the semantic and syntactic relationships. These matrices are crucial, as they encapsulate the linguistic features necessary for the models to understand and process language. To synthesize the information from these matrices, we applied a pooling operation over rows, effectively consolidating the multidimensional data into a single comprehensive vector that represents the input's semantic essence. This process is pivotal for the subsequent classification task, as it distills the essence of the input into a form that allows the models to analyze and categorize efficiently.

### *2.3. Depression Detection Pipeline*

*2.3.1. Feature Extraction*

One-dimensional Convolutional Neural Networks (1D CNNs) are increasingly acknowledged as a powerful option for categorizing text. Research has shown that 1D CNNs can be used successfully, leveraging their ability to identify features that are not dependent on their position. Combining 1D CNNs with different methods and models highlights their adaptability and the possibility for enhancing precision in text classification [42, 43]. Additionally, enhancing convolutional neural networks (CNNs) by integrating fully connected (FC) layers has shown to notably boost the model's ability to generalize from limited data samples [44].

Based on previous studies and the limited amount of available training data, we created a combinatorial network that consists of fully connected and CNN layers. This network is designed to be simple to prevent overfitting while being powerful for detecting depression. To elaborate, the feature extraction consists of a series of fully connected layers (FC) and two successive 1D convolutional neural networks (CNNs). In the first place, a block of FC layers is used to increase the dimensionality of the pooled vector in order to capture more complex patterns. The concatenated vector of the preceding block is then applied to two 1D CNN layers with various kernel sizes to identify local patterns and dependencies in the text.

To introduce some regularization we added layer normalization, a GELU activation function [45], and a dropout layer after each fully connected layer. For the CNN layers, we used batch normalization to normalize the feature maps across the batch dimension. We also used different pooling strategies for the first and second CNN layers: max pooling for the first one and average pooling for the second one. This way, the network could capture both the most salient and the average features from the convolutional outputs.

*2.3.2. Classification*

The information obtained from the feature extraction section passes through a classification part to specify the presence of depression in the participant. This part comprises two FC layers that reduce the dimensionality of the vector and produce a score for each possible label. Here also, the same regularization strategy used for the first FC block is used.

### *2.4. Depression Lexicon Construction*

*2.4.1. Tokenizer*

We employed a different tokenization method than BERT-based models from the NLTK library, called “tweet tokenizer”. This is an advanced tokenizer that can handle sentences with various symbols, such as hashtags and abbreviations. It allows us to preserve the original structure and meaning of the texts, while splitting them into smaller units for analysis.

Two methodologies were used to process the textual data and compute the word frequency. Initially, the TweetTokenizer class from the NLTK library was used to partition the text into individual tokens. The informal language and unique symbols are handled in this class. Furthermore, the Tfidfvectorizer from the Sklearn package was used to convert the tokens into numerical vectors. The present class is designed to calculate the term frequency-inverse document frequency (TF-IDF) values for every token. In our study, we only used the term frequency component of the TF-IDF value, which signifies the frequency of occurrence of a token inside a tweet. In addition, a predetermined inventory of stop words was included into our analysis by leveraging the NLTK library, with the exception of specific pronouns related to the depression alerting words.

*2.4.2. Word Score*

TF-IDF is a score that measures how important a word is in a document or a collection of documents. It is based on the term frequency (how often the word appears in the document) and the inverse document frequency (how rare the word is across the documents) and is expressed as follows:

$$(tf - idf)_{x,y} = tf_{x,y}.\log\left(\frac{N}{df_x}\right) \quad (2)$$

in which $tf_{x,y}, N, df_x$ are repetition of x in y, total number of documents, and number of documents where x is included respectively. A word with high term frequency and low document frequency gets a higher score.

The TF-IDF technique, which gives high weights to terms that are uncommon in a corpus but common in a text, is not appropriate for the depression lexicon since it may capture words that are idiosyncratic to a particular group of patients, reflecting their personal or cultural backgrounds. However, the depression lexicon aims to be generalizable to all people with depression, regardless of their individual or contextual factors.

We propose a new formula called "Word Score" (WS) that incorporates two factors: 1) weigh words that are frequent not only in one document but also in all transcripts, since this reflects the common patterns of speech among depressed patients. Adopting this approach, the TF-IDF problem is resolved; and 2) adjust the weights in accordance with the PHQ8 score to pay more attention to words used by those who are severely depressed. This is because their linguistic pattern is more valuable for us compared to individuals with slight degrees of depression.
The following equation shows our suggested WS:

$$WS = \frac{\sum(positive\ RPHQ\ score * term\ frequency)}{\sum(positive\ RPHQ\ scores)} + \frac{\sum(negative\ RPHQ\ score * term\ frequency)}{-\sum(negative\ RPHQ\ scores)} \quad (3)$$

On the right-hand side of the equation, the positive score measures how much the word is associated with a positive mood, while the negative score measures how much the word is associated with a negative mood. In both parts, we sum over all the documents that contain the word and divide by the total RPHQ score of all documents. The RPHQ score is a modified version of the PHQ score and is presented in Table 1. We use the RPHQ score to make it negative for depressed people and positive for non-depressed ones, so that higher magnitudes indicate the importance of that word. In this way, words with a negative WS go under the category of depressed words, while words with a positive WS are the opposite. The absolute magnitude of the WS also reflects the significance of the word for the class. For example, a word with WS = -4 is more important than a word with WS = -1 for depressed people.

**Table.1.** PHQ score with their associated RPHQ score for both depressed and non-depressed participants

| | | | | | | | | | | | | | | | |
|---|---|---|---|---|---|---|---|---|---|---|---|---|---|---|---|
| depressed | PHQ score | 10 | 11 | 12 | 13 | 14 | 15 | 16 | 17 | 18 | 19 | 20 | 21 | 22 | 23 |
| | RPHQ score | -1 | -2 | -3 | -4 | -5 | -6 | -7 | -8 | -9 | -10 | -11 | -12 | -13 | -14 |
| non depressed | PHQ score | 0 | 1 | 2 | 3 | 4 | 5 | 6 | 7 | 8 | 9 | | | | |
| | RPHQ score | 10 | 9 | 8 | 7 | 6 | 5 | 4 | 3 | 2 | 1 | | | | |

We used WS on the entire DAIC-WOZ data set (train, test, and validation) to extract the candidate words for the depression lexicon.

*2.4.3. Alerting Words*

In our study, we performed a thorough examination of existing literature on the relationship between depression and language use. Drawing from previous studies [38, 39, 46] that identified specific language patterns associated with depression on social media, we compiled a comprehensive list of words, which we refer to as "alerting words," that can indicate depressive symptoms. We presented this list in Table 2, categorizing the words based on an adapted and modified version of M. De Choudhury's [39] original classification. Each row in the table corresponds to a specific category along with its related words, providing a comprehensive and specific word domain for depression within the scope of our research.

**Table.2.** List of alerting words for depression based on social media surveys.

| Main Groups | Words |
|---|---|
| **(Physical/ Mental) symptoms** | depression, anxiety, delusions, adhd, overweight, insomnia, drowsiness, no appetite, dizziness, nausea, lack of sleep, seizures, addiction, mood swings, dysfunction, blurred, irritability, headache, fatigue, imbalancement, nervousness, psychosis, drowsy, eating habits, exhaustion, pain, sadness, stress, tiredness, sensitive, melancholy, oblivion (forgetfulness), cut wrist, no focus, isolation, lack of interest, moody, instability, low self-esteem, restless(ness) |
| **Treatment** | medication, side-effects, doctor, doses, prescribed, therapy, inhibitor, stimulant, antidepressant, patients, neurotransmitters, prescriptions, psychotherapy, diagnosis, clinical, pills, chemical, counteract, toxicity, hospitalization, sedative, drugs, serotonin, amphetamine, maprotiline, nefazodone |
| **Negative (words/ thoughts)** | severe, suicidal, helpless(ness), lonely(ness), hate, fear, idiot, despair, sorry, collapse, death, give up, low, leave, danger, lost, shadow, destroy, crash, dark, negative, rubbish, haze, suffer, uneasiness, worthless, kill, withdrawal, frustrated, unhappy, broken, disturbed |
| **Religious involvement** | Jesus, religion, bible, church, lord, heaven, hell, guilt, god |
| **Pronouns** | high rate of first person pronouns (I, myself, etc.) |

We considered the last two rows ('Religious involvement' and 'Pronouns') only for comparison with the list of candidate words and are not taken into account for the final depression lexicon. This is done by checking if the words in these rows are present in them, and if so, how high their word scores are.

*2.4.4. Cosine Similarity*

Cosine similarity is a measure of how similar two vectors are in terms of their orientation. It is based on the cosine distance, which is the angle between the vectors. The smaller the angle, the higher the similarity. Cosine similarity is often used in natural language processing for finding the most similar and relevant words or phrases to a document [46, 47], based on their word embeddings or vector representations.

We utilized a DistilBERT model to encode the first three categories (rows) of the table and candidate words as a vector. Next, we calculate the cosine similarity between them, which gives us a measure of how well the word fits the category. This allows us to obtain more accurate and comprehensive results than using a fixed lexicon. Finally, we sort the candidate words by their similarity scores and add them to the table accordingly.

## 3. Results and Discussions

In this section, we present and discuss various aspects of our results. We used the Google Collaboratory platform with the python programming language to build different models and run multiple experiments. We also used Pytorch, Hugging Face models, and Sklearn package to program the models evaluated in this study.

One of the goals of our project is to ensure the reproducibility of our results. Therefore, we have stored the model weights, the encoded representations of the summarized texts, and detailed table of hyper parameters for our model configuration as well as ML ones in a public repository [48]. Other researchers can access and use them for further analysis or comparison.

### *3.1. Training Setup*

In this study, we utilized a summarization approach outlined in section 2.2, and employed the KeyBERT python package for this purpose. KeyBERT provides a variety of hyperparameters that can be adjusted to impact the quality of the generated summaries. We tested various values to determine the most optimal settings for our model. One of the hyperparameters we specifically focused on is the size of the n-gram, which was selected in the range of 6 to 9 to assess the impact of summarized sentence length on the classification results. Additionally, based on recommendation from M. Grootendorst [32], we set the value of $\lambda$ to 0.7 to promote greater diversification in the final list of extracted keywords.

In section 2, we explained that BERT-based models are utilized to convert the summarized texts into numerical vectors. Due to the limited size of our dataset, we did not fine-tune the models and instead used pretrained ones known as "base-uncased," which have been trained on Wikipedia and BookCorpus texts.

In order to train our proposed depression detection pipeline, which includes feature extraction and classification parts, we utilized Focal Loss and the AdamW optimizer [49] along with pooled numerical vectors from BERT-based models as input for 100 epochs. As mentioned in section one, we employed Focal Loss to address data imbalance and improve model accuracy by focusing on challenging examples. The key hyperparameter in this function is the gamma parameter, which determines how quickly the loss decreases as the predicted probability approaches the actual class. When gamma is set to zero, Focal Loss is similar to categorical cross-entropy. Increasing gamma enhances the influence of the modulating factor, assisting in handling imbalanced classes. Following the recommended value in their paper [37], we set it to two. To give priority to the minority class, we also set the weights at [1.4, 3.3] based on the class distribution in our dataset. Additionally, the use of the AdamW optimizer ensured adaptive learning rate during the optimization process. The weight decay and learning rate of this optimizer were set at 1e-2 and 1e-3, respectively.

To assess the effectiveness of our proposed pipeline for detecting depression, we have tested various advanced machine learning (ML) models including logistic regression, SVM, and XGBoost [50] and then compared the results of these tests. It is important to highlight that we fine-tuned all of these ML models using the GridSearch class from the Sklearn library to obtain the best results. The chosen hyperparameters for all cases are included in the Table.7 (appendix section).

### *3.2. Classification Results on the development set*

We used the same experimental protocols as the DAIC-WOZ dataset proposed for our study. These protocols include calculating precision, recall and F1-score as the evaluation metrics. These metrics are widely used for comparing models on unbalanced datasets.

The tables below compare the depression detection scores of various BERT-based models and our classification framework with other ML algorithms for different n-grams. The highest scores in each table are highlighted.

**Table.3.** Classification results using ALBERT model

| ngram | method | f1 | recall | precision |
|---|---|---|---|---|
| 6 | ours | 0.69 | **0.92** | 0.55 |
| | xgboost | 0.6 | 0.5 | 0.75 |
| | logistic regression | 0.64 | 0.58 | 0.7 |
| | svm | 0.53 | 0.42 | 0.71 |
| 7 | ours | 0.71 | 0.83 | 0.62 |
| | xgboost | 0.56 | 0.42 | 0.83 |
| | logistic regression | 0.69 | **0.92** | 0.55 |
| | svm | 0.62 | 0.67 | 0.57 |
| 8 | ours | 0.73 | 0.67 | 0.8 |
| | xgboost | 0.5 | 0.42 | 0.62 |
| | logistic regression | **0.77** | 0.83 | 0.71 |
| | svm | 0.64 | 0.58 | 0.7 |
| 9 | ours | 0.7 | 0.67 | 0.73 |
| | xgboost | 0.59 | 0.42 | **1** |
| | logistic regression | 0.67 | 0.58 | 0.78 |
| | svm | 0.64 | 0.75 | 0.56 |

**Table.4.** Classification results using DistilBERT model

| ngram | method | f1 | recall | precision |
|---|---|---|---|---|
| | ours | 0.7 | 0.67 | 0.73 |
| | xgboost | 0.5 | 0.33 | **1** |
| 6 | logistic regression | 0.53 | 0.42 | 0.71 |
| | svm | 0.53 | 0.42 | 0.71 |
| 7 | ours | **0.8** | 0.83 | 0.77 |
| | xgboost | 0.53 | 0.42 | 0.71 |
| | logistic regression | 0.70 | 0.67 | 0.73 |
| | svm | 0.72 | 0.75 | 0.69 |
| 8 | ours | 0.69 | **0.92** | 0.55 |
| | xgboost | 0.48 | 0.42 | 0.56 |
| | logistic regression | 0.67 | 0.67 | 0.67 |
| | svm | 0.56 | 0.58 | 0.54 |
| 9 | ours | 0.73 | 0.67 | 0.8 |
| | xgboost | 0.5 | 0.42 | 0.62 |
| | logistic regression | 0.67 | 0.58 | 0.78 |
| | svm | 0.69 | 0.75 | 0.64 |

**Table.5.** Classification results using BERT model

| **ngram** | **method** | **f1** | **recall** | **precision** |
|---|---|---|---|---|
| 6 | ours | 0.71 | 0.83 | 0.62 |
| | xgboost | 0.67 | 0.5 | **1** |
| | logistic regression | 0.43 | 0.42 | 0.45 |
| | svm | 0.5 | 0.5 | 0.5 |
| 7 | ours | **0.81** | **0.92** | 0.73 |
| | xgboost | 0.5 | 0.42 | 0.62 |
| | logistic regression | 0.6 | 0.75 | 0.5 |
| | svm | 0.61 | 0.58 | 0.64 |
| 8 | ours | 0.61 | **0.92** | 0.46 |
| | xgboost | 0.29 | 0.17 | **1** |
| | logistic regression | 0.56 | 0.58 | 0.54 |
| | svm | 0.45 | 0.42 | 0.5 |
| 9 | ours | 0.8 | 0.83 | 0.77 |
| | xgboost | 0.6 | 0.5 | 0.75 |
| | logistic regression | 0.77 | 0.83 | 0.71 |
| | svm | 0.64 | 0.67 | 0.62 |

Based on the preceding tables, our depression detection pipeline demonstrated better performance, surpassing all three machine learning algorithms in most cases. It achieved an f1-score of 0.81 and a recall score of 0.92. Following our model, logistic regression emerged as the second-best performer. The strength of our model lies in expanding the feature space, making it more informative and enabling the model to learn internal intricacies. Additionally, its ability to capture local dependencies through 1D CNN networks which is essential in text data. On the other hand, logistic regression outperformed other ML algorithms due to the small size of the data and binary nature of the problem.

Generally, an n value of 7 produced superior results compared to other n-grams. This demonstrates that a seven-token sentence can encapsulate the essence of the entire document and can be considered as a starting point for future summarization studies.

We achieved our best result with the BERT model, but other models also showed promising results. Therefore, even a lightweight model like ALBERT can be used for this purpose in future research.

One of the notable strengths of our approach is its high recall score, or in other words, its ability to minimize false negatives. This is particularly crucial in medical investigations, where accurately diagnosing the condition is more important than avoiding false positives. Figure 4 shows the confusion matrix that depicts our best result, giving a comprehensive overview of its performance. As shown, only one individual with depression was misclassified as non-depressed.

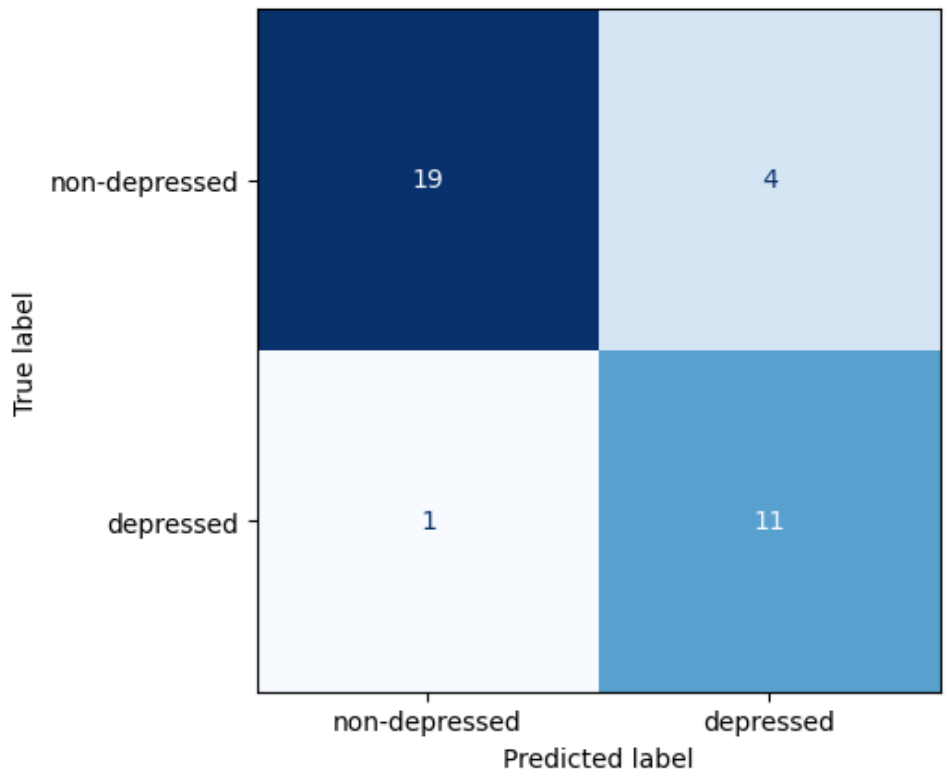

**Fig.4.** Confusion matrix of our best result

### *3.3. Performance Comparison on the development set*

Table 6 presents the best performance of our proposed pipeline (last row) compared to all the state-of-the-art results on the development set from earlier research, listed in a chronological order.

**Table.6.** Classification results of all previous studies on the development set

| **Ref** | **Scores on Dev** | | |
|---|---|---|---|
| | **F1** | **Pre** | **Rec** |
| [51] | 0.6 | 0.68 | 0.63 |
| [27] | 0.64 | 0.52 | 0.77 |
| [16] | 0.77 | 0.7 | 0.82 |
| [18] | 0.83 | 0.83 | 0.83 |
| [21] | 0.81 | 0.82 | 0.8 |
| [20] | 0.6 | _ | _ |
| [13] | 0.66 | 1 | 0.5 |
| **ours** | **0.81** | **0.73** | **0.92** |

As previously mentioned, our model excels in reducing the number of false negative cases. The table clearly shows that our depression detection pipeline achieved the highest recall score among all studies.

In the table, it is seen that the work done by L. Lin [18] achieved the highest F1 score on the validation set. They used a data resampling method to ensure an equal number of text samples in each class. However, data resampling can introduce noise to the dataset, which is especially problematic for text data. Y. Zhang [21] assigned scores to important terms in the texts, but this approach may not be easily applicable to other datasets and real-world problems. Apart from these drawbacks, both of these works required direct human intervention. In contrast, our pipeline is fully automated and does not involve any form of data augmentation or scoring system, yet it still produces satisfactory results.

### *3.4. Performance Comparison on the test set*

There are only a few papers that assessed their model on the test set. The winner of the AVEC 2016 competition and A. Mallol-Ragolta [20] achieved an f1-score of 0.57 and 0.63 on the test set, respectively. Our proposed pipeline successfully achieved an f1-score of 0.67 on the test set, surpassing previous results by a meaningful margin.

### *3.5. Word Score Analysis*

Figure.5. displays the top thirty terms with the highest Word Score (WS) in the DAIC-WOZ dataset. Consistent with the alerting words noted in social media analysis, it is apparent that individuals dealing with depression often use first-person pronouns ("I," "me," and "myself") with significant frequency. Specifically, the pronoun "I" stands out with the highest absolute word score among all words with a negative WS.

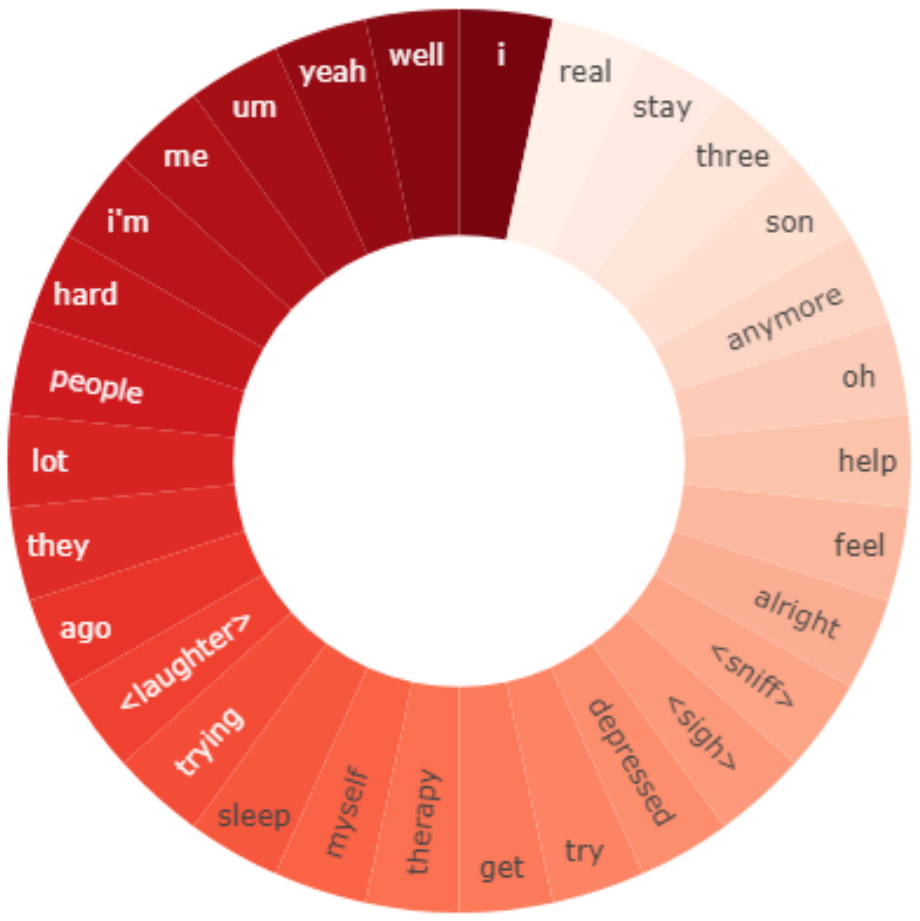

**Fig. 5.** The words with lowest negative WS (highest absolute WS for negative words). The significance of words is shown by the intensity of color.

Based on our findings, we observed that certain religious terms like "religion," "hell," and "Jesus" had negative word scores, whereas "bible," "church," "lord," "heaven," "guilt," and "God" had positive word scores. As a result, the use of religious terms may not be a reliable indicator for detecting depression in this particular dataset.

### *3.6. Depression Lexicon*

In this section, we present the extracted keywords of the DAIC-WOZ dataset using our suggested approach. We calculated the Word Score (WS) for the entire dataset and selected a subset of 2000 words with the lowest values as potential indicators of depression. Subsequently, we computed the cosine similarity between each individual word and every category listed in Table 2. To achieve this, we summed the cosine similarity scores between the given term and all other words within the same category. This allowed us to assign each word to its most relevant group. It's important to note that some terms were found to be present in multiple categories. To determine their appropriate categorization, we assigned them to the groups that showed the highest cosine similarity scores. The following three word clouds show the results for the top twenty terms within each respective category, with the significance of words indicated by their size.

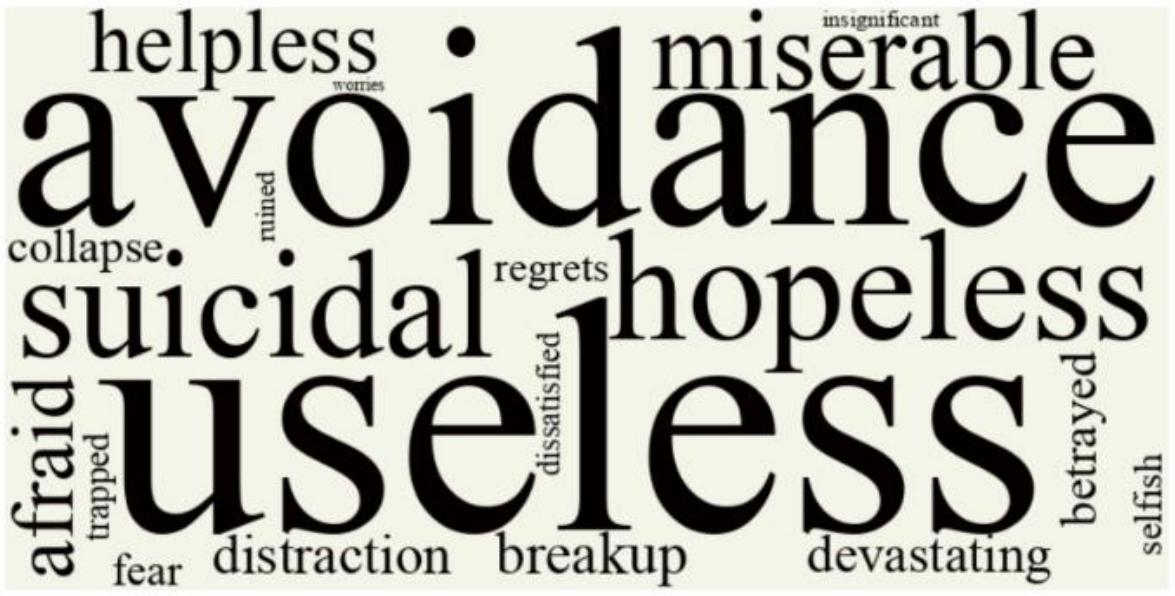

**Fig. 6.** Depression lexicon for the category of negative words.

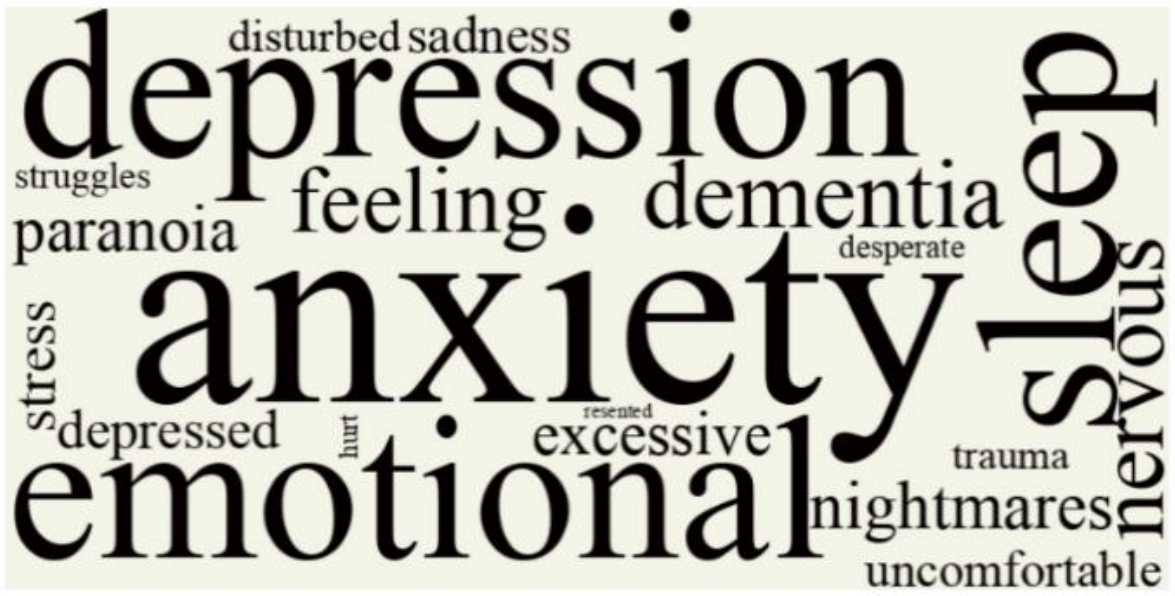

**Fig. 7.** Depression lexicon for the category of physical and mental symptoms.

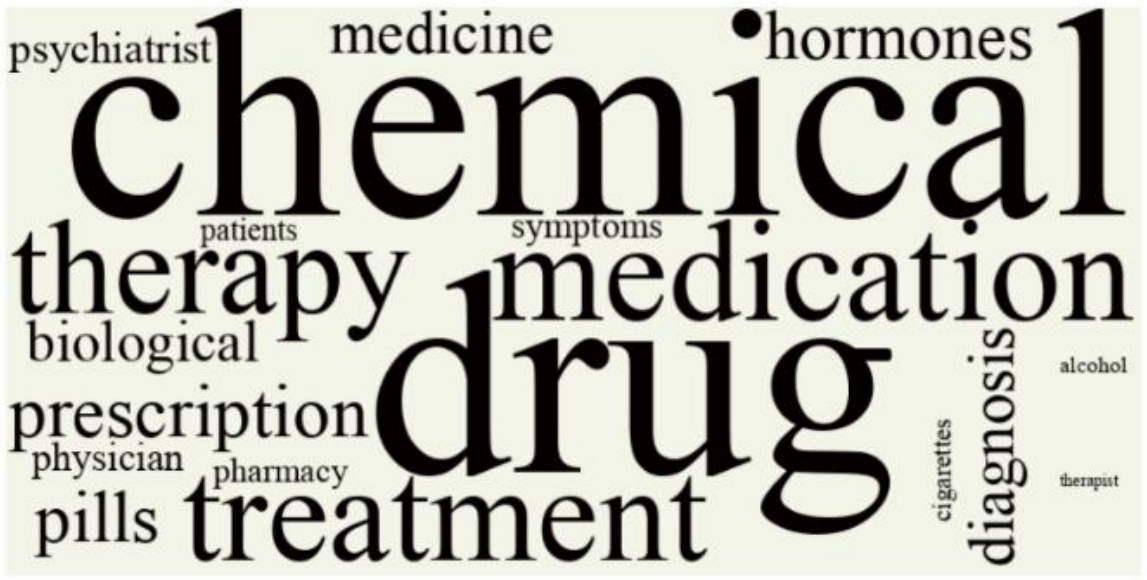

**Fig. 8.** Depression lexicon for the category of treatment.

In previous studies on online platforms, we have noticed some overlap in the words indicating depression compared to our findings. However, we have also significantly expanded the vocabulary related to depression. Our work holds more merit because our data consisted of longer texts rather than brief tweets or posts online. Additionally, interviews specifically for the diagnosis of depression are clinically more reliable and provide more relevant information about depression compared to online posts.

Based on our findings, it seems that using KeyBERT for summarization can be a promising and reliable approach. The fact that 75% of the words from the depressive lexicon were included in the summary texts provided by KeyBERT indicates that it is highly proficient in identifying and extracting key words and their associated phrases. This makes KeyBERT a valuable tool for accurate summarization, and it could significantly benefit various applications that require efficient text summarization.

### *3.7. Conclusion and Future Works*

Text summarization is a technique that reduces the length of textual documents to make their analysis easier. It is a valuable tool for various domains dealing with large amounts of text, such as journalism, education, research, and business.

This study presents a new method for detecting depression using KeyBERT, which is a BERT-based approach for summarizing text. The method is tested on the DAIC-WOZ dataset, which includes interviews with individuals diagnosed with depression and those without depressive symptoms. Unlike other methods that rely on text augmentation or manual feature selection, our method is fully automated and can identify the most relevant phrases for identifying depression. Our classification results show that our method is competitive compared to other state-of-the-art methods. To enhance the credibility of our method, we create a depression lexicon tailored to the DAIC-WOZ dataset. We then demonstrate that most of the important terms identified by KeyBERT are aligned with the lexicon. This suggests that our method is robust and can be applied to other datasets.

Developing subject-oriented summarizers, in which the model detects key phrases according to a predefined word domain, is a promising field of research. By employing this approach, the summarizing method will intelligently condense texts of varying lengths. This involves selecting terms closely associated with the subject matter, resulting in a higher classification score. Therefore, as a direction for further work, we plan to develop a novel summarization technique that leverages our depression lexicon to generate concise and informative summaries for the given dataset as well as others related to depression.

**Appendix**

**Table.7.** Chosen hyperparameters for ML models in differenct cases

| ML Model | | n = 6 | n = 7 | n = 8 | n = 9 |
|---|---|---|---|---|---|
| Logistic Regression (class_weight = 'balanced' max_iter = 1000) | BERT | c = 1<br>solver = saga<br>tol = 0.01 | c = 0.5<br>solver = sag<br>tol = 0.01 | c = 1.5<br>solver = sag<br>tol = 1e-5 | c = 2<br>solver = liblinear<br>tol = 0.01 |
| | DistilBERT | c = 0.5<br>solver = lbfgs<br>tol = 0.01 | c = 0.5<br>solver = saga<br>tol = 0.001 | c = 0.5<br>solver = liblinear<br>tol = 0.01 | c = 0.5<br>solver = liblinear<br>tol = 0.01 |
| | ALBERT | c = 1.5<br>solver = sag<br>tol = 0.01 | c = 0.5<br>solver = sag<br>tol = 0.01 | c = 0.5<br>solver = sag<br>tol = 0.0001 | c = 1<br>solver = newton<br>tol = 0.001 |
| SVM (class_weight = 'balanced' max_iter = 1000 | BERT | c = 1.5<br>kernel = sigmoid<br>tol = 0.01 | c = 1<br>kernel = sigmoid<br>tol = 0.01 | c = 0.5<br>kernel = linear<br>tol = 0.01 | c = 0.5<br>kernel = sigmoid<br>tol = 0.01 |
| | DistilBERT | c = 1.5<br>kernel = sigmoid<br>tol = 0.01 | c = 1.5<br>kernel = sigmoid<br>tol = 0.01 | c = 0.5<br>kernel = sigmoid<br>tol = 0.01 | c = 0.5<br>kernel = sigmoid<br>tol = 0.01 |
| | ALBERT | c = 0.5<br>kernel = linear<br>tol = 0.01 | c = 3<br>kernel = sigmoid<br>tol = 0.01 | c = 0.5<br>kernel = linear<br>tol = 0.01 | c = 0.5<br>kernel = sigmoid<br>tol = 0.01 |
| XGBoost (n_estimators = 100 booster = gbtree) | BERT | gamma = 0<br>lr = 0.4<br>max_depth = 3<br>min_child_weight = 3<br>reg_alpha = 0.08<br>reg_lambda = 0.5 | gamma = 0<br>lr = 0.1<br>max_depth = 3<br>min_child_weight = 5<br>reg_alpha = 0.08<br>reg_lambda = 0.4 | gamma = 1<br>lr = 0.1<br>max_depth = 4<br>min_child_weight = 3<br>reg_alpha = 0.08<br>reg_lambda = 0.4 | gamma = 1<br>lr = 0.4<br>max_depth = 3<br>min_child_weight = 4<br>reg_alpha = 0.08<br>reg_lambda = 0.4 |
| | DistilBERT | gamma = 0<br>lr = 0.2<br>max_depth = 3<br>min_child_weight = 3<br>reg_alpha = 0.08<br>reg_lambda = 0.4 | gamma = 0<br>lr = 0.2<br>max_depth = 3<br>min_child_weight = 4<br>reg_alpha = 0.08<br>reg_lambda = 0.4 | gamma = 1<br>lr = 0.3<br>max_depth = 3<br>min_child_weight = 4<br>reg_alpha = 0.08<br>reg_lambda = 0.4 | gamma = 0<br>lr = 0.2<br>max_depth = 3<br>min_child_weight = 4<br>reg_alpha = 0.08<br>reg_lambda = 0.4 |
| | ALBERT | gamma = 0<br>lr = 0.4<br>max_depth = 3<br>min_child_weight = 3<br>reg_alpha = 0.08<br>reg_lambda = 0.5 | gamma = 0<br>lr = 0.3<br>max_depth = 3<br>min_child_weight = 3<br>reg_alpha = 0.08<br>reg_lambda = 0.4 | gamma = 0<br>lr = 0.2<br>max_depth = 3<br>min_child_weight = 3<br>reg_alpha = 0.08<br>reg_lambda = 0.4 | gamma = 0<br>lr = 0.2<br>max_depth = 3<br>min_child_weight = 3<br>reg_alpha = 0.1<br>reg_lambda = 0.4 |